\documentclass[12pt,a4paper]{article}
\usepackage{nips14submit_e,times}
\usepackage{hyperref}
\usepackage{url}
\usepackage{listings}
\usepackage[linesnumbered,boxed,ruled,commentsnumbered]{algorithm2e}
\usepackage[top=3cm,bottom=2cm,left=2cm,right=2cm]{geometry} 
\usepackage{amsmath} 
\usepackage{amsthm}
\usepackage{longtable} 
\usepackage{graphicx} 
\usepackage{tikz} 
\usepackage{cite}
\usepackage{listings}
\usepackage{amsfonts}
\usepackage{subfigure}
\usepackage{float}
\usepackage{appendix}
\usepackage[linesnumbered,boxed,ruled,commentsnumbered]{algorithm2e}

\renewenvironment{abstract}{\vskip.075in\centerline{\large\bf
Abstract}\vspace{0.5ex}\begin{quote}}{\par\end{quote}\vskip 1ex}

\nipsfinaltrue
\hypersetup{CJKbookmarks=true}

\title{Second-order Information in First-order Optimization Methods}

\author{
{\large Yuzheng Hu${}^{{}_{\footnotemark}}$\qquad Licong Lin${}^*$\qquad Shange Tang${}^*$}\\
Peking University, School of Mathematical Science\\
Beijing,China\\
\{1700010613,llc2000,sangertang\}@pku.edu.cn
}

\begin{document}
\maketitle
\begin{abstract}
First-order optimization methods play a central role in modern deep learning. There are various types of first-order methods, such as GD, SGD, Momentum, NAG, Adagrad, RMSprop, Adam, etc. An interesting fact is that some algorithms are more favored in practice (SGD and Adam, for example) and experiments have revealed that these algorithms are indeed significantly better than others. Although there are many heuristic explanations trying to interpret this phenomenon, theoretical guarantees are still lacking.

In this paper, we aim to uncover part of the mystery from a novel perspective. We claim that some first-order methods, though Hessian-free, are actually utilizing the second-order information of the loss function while training. More specifically, we demonstrate that a powerful method, Nesterov Accelerated Gradient, explores second order information by making use of the difference of past and current gradients and thus accelerates the training. Furthermore, we rigorously prove that adaptive methods such as Adagrad and Adam can actually be regarded as relaxations of natural gradient descent---a well-known second-order technique in computation statistics. Based on this observation, we design a new algorithm, AdaSqrt, which has better performance on MNIST and CIFAR10 even compared to the most popular method, Adam. This surprising result casts doubt on the convention view that the square root in the denominator of the adaptive method is crucial for training and makes us rethink the core of these algorithms.

We hope that our work can shed light on the importance of second-order information or generally, higher-order information, and encourage the design of stronger optimization algorithms in the future.
\end{abstract}

\section{Introduction}
In modern machine learning field, with the explosive growth of computing power and the resurgence of deep learning, people are beginning to design new models with extremely high capacity, some of which reaching millions of parameters (ResNet). Training such models require the use of gradients or Hessians, however, it is difficult to compute and store the Hessian in high dimensionality. Thus, second-order methods such as Newton method do not apply to deep learning and first-order methods naturally become the center of modern optimization.\footnotetext[1]{Co-first authors contributed equally to this paper. }

Over the past decades, a number of first-order methods have been brought up and widely used in practice. The primitive method is simply gradient descent, which updates the parameters of the model in the opposite direction of the gradient of the objective function. The idea of gradient descent is quite naive and can be viewed as a greedy algorithm: when the learning rate is fixed, gradient descent just chooses a direction that forces the objective function to decrease most at current position. All other gradient-based algorithms are founded upon this simple intuition and they make small but delicate modifications upon Gradient Descent. When adding noise to the gradients, we immediately have SGD. When enforcing the gradients to be stable along the trajectory, we get SGD plus momentum and Nesterov Accelerated Gradient. When considering the choice of different learning rate at different stage of training, we have a series of important work---all referred to as adaptive method, for example, Adagrad and Adam.

Although these variants of gradient methods look reasonable at first glance and have decent performance in practice, several questions remain largely unknown even to theoretists. For example, why does SGD and Adam constantly perform better in practice? Another intriguing question is: how does Momentum help with training and why Nesterov Accelerate Gradient is regarded as the��strongest first-order method��? On top of that, the key motivation of this work is based on a simple observation that is often dismissed easily but somehow inspires our curiosity: why do adaptive methods use the square root of second moment of past gradients instead of directly using the second moment?

There are indeed quite a lot interpretations concerning the pros and cons of these modified versions of GD and partially answer the questions raised above. For example, it is claimed that SGD works good because of the injected noise that potentially leads the algorithm to escape from saddle points or bad local minimum. Momentum are believed to accelerate the training process since oscillation is neutralized along the optimization trajectory. Also, adaptive methods are popular because they set different learning rate for different coordinates, which make perfect sense for models such as NLP. Nevertheless, these arguments are mostly vague and debatable and rigorous theoretical guarantees are eagerly needed. All these heuristic accounts of the advantage of each algorithm might be adequate for programmers, but it is far from satisfactory for mathematicians or statisticians.

In this paper, we aim to uncover part of these mysteries through a novel perspective: second-order information. More precisely, we believe that some of the first-order methods are implicitly exploring second-order information. It is well known that higher-order method can accelerate the rate of convergence, thus making use of higher-order information about the objective function will definitely help training and generalization. We consider two types of first-order methods: momentum and adaptive method, among which we choose three typical algorithms: Nesterov Accelerated Gradient and Adagrad \& Adam to analyze. For starters, we consider Nesterov Accelerated Gradient and rigorously prove that this algorithm explores second order information by making use of the difference of past and current gradients. Furthermore, we rigorously prove that adaptive method (Adagrad, Adam) can actually be regarded as relaxations of natural gradient descent---a well-known second-order method in computation statistics. Based on this observation, we design a new algorithm, AdaSqrt, which outperforms Adam on MNIST and CIFAR10. This result casts doubt on the tradition view that the square root in the denominator of the adaptive method is crucial and training without it will lead to terrible performance.

We believe that our contributions lie in at least four aspects:
\begin{enumerate}
\item	We shed light on the importance of second-order information for training. In spite of the huge success of SGD, we still believe that going beyond gradient is a promising approach with huge practical value that deserves more attention in the future.
\item We propose a new algorithm, AdaSqrt, which outperforms the most popular method for training neural networks, Adam, on MNIST and CIFAR10. For lacking of computing resources, we do not have enough time to carry out our experiments on ImageNet but we plan to do this work in Computing Center after the course.
\item The success of our new algorithm demonstrates that the convention view: the square root of denominator in the adaptive method is crucial for training, is likely to be untrue. This result inspires us to rethink the essence of these powerful algorithms: we believe that as long as the algorithm seeks to explore second-order information of the objective function, then proper scaling of the learning rate alone can guarantee good training and generalization performance. To the best of our knowledge, this is the first paper that consider the necessity of square root in adaptive methods.
\item Our new perspective of analyzing first-order optimization method through second-order information potentially leads to the design of more powerful algorithms that seek to explore higher-order information in future.
\end{enumerate}
\qquad The remaining of this paper unfolds as follows. In Section2 we will summary related works. In section3 we will analyze Nesterov Accelerated Gradient and reveal the second-order essence of this algorithm. In Section4 and 5 we will dig into adaptive method, uncover the relationship between Adam(Adagrad) and NGD and state our new algorithm AdaSqrt with a convergence theorem. In Section6 we will present our experiment results. In Section7 we make conclusions and propose some future works.

\section{Related Works}
\subsection{First-order optimization methods}
First order optimization methods have been studied extensively since the resurgence of deep learning \cite{1606.04838}\cite{1609.04747}\cite{3}. The most classical methods are Gradient Descent and Stochastic Gradient Descent\cite{4}. Gradient Descent works perfect in traditional machine learning models, especially when the objective function has some nice properties, such as convex or strongly-convex. For deep neural networks, since the loss surface is highly-nonconvex, it is intuitively that Gradient Descent will get stuck in saddle points or bad local minimum and thus result in bad training and generalization performance. Recently, however, it is reported that gradient descent with Gaussian initialization can achieve zero loss in ultra-wide neural networks\cite{1810.02054}. For Stochastic Gradient Descent, a common understanding of this algorithm is that it has the power of implicit regularization---that is, without explicitly penalizing the norm of the weights, Stochastic Gradient Descent naturally finds a solution with low "complexity"\cite{1811.00659}. However, theoretical understanding of SGD remains largely unknown. There is little theoretical guarantee that SGD will converge to zero on a neural network with moderate width, nor are there sufficient explanations for the good generalization performance of SGD, aside from its highly chaotic trajectory.

In this paper, however, we leave alone the huge empirical success of SGD and mainly focus on adaptive methods \cite{1212.5071}\cite{Coursera}\cite{1412.6980}\cite{1502.04390}\cite{Duchi}
and Nesterov Accelerated Gradient Algorithm\cite{12}\cite{1503.01243} \cite{14}. Adaptive methods are the ensemble of a series of algorithms---these algorithms all use adaptive learning rates rather than a fixed one. Although people manually adjust the learning rate of SGD or GD in practice, adaptive methods have the power of choosing learning automatically. More specifically, Adagrad\cite{Duchi} uses the inverse square root of second moment of past gradients as a scaling factor. Noticing that the inverse square root of second moment potentially points to zero as the training time approaches infinity, RMSprop\cite{Coursera} alleviates this risk by adjusting the coefficients of past gradients. Adam\cite{1412.6980} makes addition modification on RMSprop by using momentum on past gradients. Among these three algorithms, Adam works best in practice and has become the most popular optimization method since its emergence. In recent years, more adaptive methods were brought up by researchers in hope of defeating Adam. For example, some observed that the learning rate of Adam oscillates in the mid-to-late phase of training, thus they proposed to clip the stepsize in this stage to stabilize the training procedure, which yields Adabound\cite{1902.09843}.

One should bear in mind that though some adaptive methods employ the idea of using past gradients, the core of these algorithms still lie in "adaptive" ---that is, these algorithms explicitly explore the geometry property of loss surface and make use of it when choosing new learning rates. On the other hand, momentum\cite{16}\cite{17}\cite{18} alone already forms a line of research. The intuition behind this method is to avoid the fluctuation of changing gradients, since the vanilla SGD potentially leads to bouncing trajectory within a local valley, thus slowing down the training process. Momentum naturally enforces the training trajectory to be smooth, as the new update directions involved the history of past gradient. Nesterov Accelerated Gradient is a splendid adaption of momentum. It not only makes use of past gradients, but also utilizes the envision of future. More specifically, NAG estimates the future location by using information of past gradients, and directly uses the foreseen gradient to update the weights. This small modification leads to considerable improvement in certain frameworks\cite{19}.

\subsection{Natural Gradient Descent}
Natural Gradient Descent can be traced back to Amari's work in 1985 \cite{20}, where the notation of information geometry was first put forward. Amari combined statistics with differential geometry, hence methods in geometry can be applied to study the properties of distributions. In fact, a distribution family can be viewed as a Riemannian manifold, whose metric is given by the Fisher Information Matrix\cite{21}. Under the Fisher metric, we can also solve optimization problems. Therefore, given the probabilistic interpretation of any statistical model, gradient methods can be applied to it, where the "gradient" is taken over Fisher metric. Here, the gradient is called "Natural Gradient", which can be rigorously defined as the inverse of Fisher Information Matrix times the ordinary gradient.

Natural Gradient Descent, or NGD, is considered to be an efficient alternative to traditional gradient descent methods. Since the resurgence of deep learning, NGD has been applied to various kinds of problems related to deep learning, such as neural networks\cite{22}\cite{23}\cite{24}\cite{Park}\cite{Arnold} \cite{Martensf}, reinforcement learning\cite{Peters}, etc. NGD is thought to be better than ordinary gradient methods mainly for two reasons. The first reason is that NGD possesses nice properties such as parameterization invariance, while ordinary gradient methods highly depend on the parametrization of the model\cite{1301.3584}. To some extent, NGD is more robust. The second reason is that NGD can be viewed as a type of second order method
\cite{1412.1193}. NGD reveals information of the distribution space (manifold), which tends to make the optimization procedure quicker and smoother. However, in high dimensional case, the inverse of the Fisher Information matrix is hard to compute, so it is important to design provable algorithms that approximate the Fisher Matrix appropriately in practice.\cite{1412.1193}

There are also other optimization methods closely related to NGD: Hessian Free Optimization\cite{Martensz}, Krylov Subspace Descent \cite{Vinyals} The relationship between those methods are discussed in \cite{1301.3584}.

\section{A Quick Warmup: Momentum and Nesterov Accelerated Gradient}
\subsection{Momentum}
In this section, we focus on a branch of first-order optimization method: Momentum. Momentum was first brought up by Geoffrey Hinton in 1986 in light of the hard training of traditional algorithm. In 1999, Ning Qian combined Momentum with the most popular first-order method: Gradient Descent, resulting in a dramatic improvement in the convergence rate of training. In deep learning, Momentum updates the weights of neural network as follows:
\begin{algorithm}
\caption{Momentum}
$d_i = \beta\cdot d_{i-1}+\text{gradient}(\theta_{i-1})$\\
$\theta_i=\theta_{i-1}-\alpha\cdot d_i$
\end{algorithm}
Here, $\theta_i$ represents the weights of neural network and $d_i$ represents the update direction at time $i$. Alpha is simply the learning rate and beta signifies the decaying rate of past gradients. They are both hyperparameters belonging to $(0,1)$ that require fine-tuning.

The idea of Momentum is quite simple: by enforcing the past gradients to play a part in the decision of new update direction, the algorithm naturally induces smoothness over the trajectory. One quick example: imagine a person sliding down from a $10$-th floor building, then it would be unlikely for him to reverse the direction upon reaching $2$-nd floor. But how does smoothing trajectories help with training? Recall that the loss surface contains a huge number of ravines, i.e. areas where the surface curves much more steeply in one dimension than in another\text{twoproblems}. In such case, GD or SGD will oscillate across the slopes of the ravine while only make hesitant progress along the bottom towards the local optimum with high probability. By using Momentum, however, the oscillation within the ravine is neutralized and the algorithm rush down to the local minimum steadily with an extremely fast speed. Figure 1 and 2 show that SGD with momentum indeed converges faster than ordinary SGD method at the beginning of training.
\begin{figure}[htb]	
\begin{minipage}[c]{.5\linewidth}
\centering

\includegraphics[width=23em]{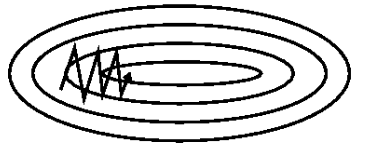}
\caption{SGD Without Momentum}
\end{minipage}
\begin{minipage}[c]{.5\linewidth}
\centering
\includegraphics[width=24em]{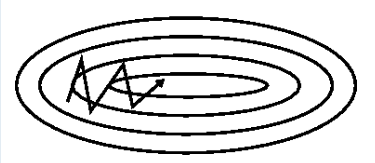}
\caption{SGD With Momentum}
\end{minipage}
\end{figure}

\subsection{Nesterov Accelerated Gradient (NAG)}

Although Momentum has achieved significant improvements upon Gradient Descent and Stochastic Gradient Descent, it is still far from satisfactory in practice. Nesterov Accelerated Gradient, however, is extremely favored by engineers and often referred to as the��best first-order optimization��. Surprisingly, this algorithm only makes a simple modification over the location of gradient upon the original Momentum:
\begin{algorithm}
\caption{Nesterov Accelerated Gradient}
$d_i = \beta\cdot d_{i-1}+\text{gradient}(\theta_{i-1}-\alpha\cdot \beta \cdot d_{i-1})$\\
$\theta_i=\theta_{i-1}-\alpha\cdot d_i$
\end{algorithm}

The high level idea of this algorithm is to apply prescience to the one-step update of parameters. Since the new direction will surely contain the momentum term by the updating rules, we can explore the future gradient in advance by cheating, or equivalently, by using momentum. In other words, NAG employs past information to predict the future. But how do we formalize this advantage in mathematics? Intuitively, since the difference between NAG and the original Momentum is in fact the difference between two gradients, it is natural for us to conjecture that NAG implicitly uses second-order information. Formally, we have the following theorem:\\
\quad\\
\textbf{Theorem 1 (Equivalence Theorem).\quad}
The original NAG algorithm is equivalent to the following Algorithm:
\begin{algorithm}
\caption{An Equivalent Form of NAG}
$d_i = \beta \cdot d_{i-1}+\text{gradient}(\theta_{i-1})+\beta \cdot(\text{gradient}(\theta_{i-1})-\text{gradient}(\theta_{i-2}))$\\
$\theta_i =\theta_{i-1}- \alpha \cdot{d_i}$
\end{algorithm}\\
\textit{Proof:} Consiter the original Momentum: $\left\{\begin{array}{l}{d_{i}=\beta d_{i-1}+g\left(\theta_{i-1}-\alpha \beta d_{i-1}\right)} \\ {\theta_{i}=\theta_{i-1}-\alpha d_{i}}\end{array}\right.$ \\
We have
\begin{align*}  \theta_{i}-\alpha \beta d_{i-1} &=\theta_{i-1}-\alpha(\beta+1) d_{i} \\ &=\theta_{i-1}-\alpha(\beta+1)\left[\beta d_{i-1}+g\left(\theta_{i-1}-\alpha\beta d_{i-1}\right)\right] \\ &=\theta_{i-1} -\alpha \beta d_{i-1}-\alpha \beta^{2} d_{i-1}-\alpha(\beta+1) g\left(\theta_{i-1}-\alpha\beta d_{i-1}\right)\hspace{5em}(*)\end{align*}
Denote
\begin{align*}  \hat{\theta}_{i} &=\theta_{i}-\alpha \beta d_{i} \\ \hat{\theta}_{i-1}&=\theta_{i-1}-\alpha \beta d_{i-1} \\ \hat{d}_{i} &=\beta^{2} d_{i-1}+(\beta+1) g\left(\theta_{i-1}-\alpha \beta d_{i-1}\right) \\ &=\beta^{2} d_{i-1}+(\beta+1) g(\hat{\theta}_{i-1}) \end{align*}
By (*) we have $\hat{\theta}_{i}=\hat{\theta}_{i-1}-\alpha \hat{d}_{i} \quad(**)$ \\
Notice that \begin{align*} \hat{d}_{i}&=\beta^{2} d_{i-1}+(\beta+1) g(\hat{\theta}_{i-1}) \\
 &=(\beta+1) g(\hat{\theta}_{i-1})+\beta^{2}\left[\beta d_{i-2}+g(\theta_{i-2}-\alpha \beta d_{i-2})\right] \\ &=(\beta+1) g(\hat{\theta}_{i-1})+\beta^{2} g(\hat{\theta}_{i-2})+\beta^{3}\left[\beta d_{i-3}+g\left(\theta_{i-3}-\alpha \beta d_{i-3}\right)\right] \\ &\quad\cdots\\ &=(\beta+1) g(\hat{\theta}_{i-1})+\sum_{j=2}^{+\infty} \beta^{j} g(\hat{\theta}_{i-j}) \end{align*}
Similarly,\begin{align*} \beta \hat{d}_{i-1} &=\beta(\beta+1) g( \hat{\theta}_{i-2})+\sum_{j=2}^{+\infty} \beta^{j+1} g(\hat{\theta}_{i-1-j}) \\ &=\beta(\beta+1) g(\hat{\theta}_{i-2})+\sum_{j=3}^{+\infty} \beta^j g(\hat{\theta}_{i-j}) \end{align*}
Thus,\quad$\hat{d}_i-\beta\hat{d}_{i-1}=(\beta+1)g(\hat{\theta}_{i-1})-\beta g(\hat{\theta}_{i-2})=g(\hat{\theta}_{i-1})+\beta [g(\hat{\theta}_{i-1})- g(\hat{\theta}_{i-2})]\hspace{3em}{(***)}$\\
\quad\\
Combining $(**)$ and $(***)$ yields the desired result.$\hspace{22em}\qedsymbol$

On the surface, NAG works better because it uses ��forward�� gradient rather than current gradient and this somewhat��cheating gradient��potentially leads to faster training. However, according to our Equivalence Theorem, this��advance��in fact plays a minor role in the empirical success of NAG and does not reveal the intrinsic property of this algorithm. As a matter of fact, we believe that the key issue here lies in the order of information: the crucial difference between NAG and the original Momentum is that NAG uses second-order information while training. More specifically, NAG adds an additional term: the difference between current gradient and formal gradient to the update rules, which can be treated as an approximation of the Hessian of the objective function. Thus, NAG can possibly reach a convergence rate of order two while the original Momentum can by no means realize such speed. This is exactly why NAG works so well in practice.

\section{\textbf{Natural Gradient Descent}}
\subsection{A Probability Interpreration of Neural Networks}
\qquad From this section, we will dig into another branch of optimization: adaptive method. For starters, we need to give an interpretation of neural networks from the probabilistic perspective and reveal the similarity between the adaptive methods and a powerful technique in computing statistics: natural gradient descent.

\subsubsection{Regression Problem}
We first consider the regression problem in machine learning. Given a training data set $ \{ \mathbf{(x_{n},t_{n})} \} $, where $n=1,2,\cdot \cdot \cdot,N$,
$\{ \mathbf{x_{n}}\} $ consider as input and $\{ \mathbf{t_{n}}\} $ consider as output, our goal is to predict the value of $ t $ for a new value of $ \mathbf{x} $. We assume that there is a statistic model
$$t=y(\mathbf{x},\mathbf{w})+\epsilon ,$$
where $\epsilon$ be a stochastic error, and $y$ be the deterministic fuction we want to derive, $\mathbf{w}$ be the parameters. For a new value of $ \mathbf{x} $, our prediction of $\mathbf{w}$ will be $y(\mathbf{x},\mathbf{w})$. \\
In the neural networks framework, $y(\mathbf{x},\mathbf{w})$ is supposed to have the form
$$y(\mathbf{x},\mathbf{w})=\sigma(W_{d} \circ \sigma(W_{d-1} \cdot \cdot \cdot \sigma(W_{1} \mathbf{x}))),$$
where $\sigma$ be the activate function(i.e., ReLU, sigmoid), and $\{ W_{i}\} $ be the weight matrices, which can be viewed as parameters of $y(\mathbf{x},\mathbf{w})$.
We can make a further assumption to specify the problem. If we assume that the stochastic error is a Gaussian Noise with zero mean and precision(inverse variance) $\beta$, then we can write
$$p(t|\mathbf{x},\mathbf{w},\beta)=\mathcal{N} (t|y(\mathbf{x},\mathbf{w}),\beta^{-1}).$$
\qquad Making the assumption that the data points are drawn independently from the distribution above, we obtained the likelihood fuction
$$p(\mathbf{t}|\mathbf{X},\mathbf{w},\beta)=\Pi_{i=1}^{N}\mathcal{N} (t_{i}|y(\mathbf{x_{i}},\mathbf{w}),\beta^{-1})$$
where $\mathbf{t}=(t_{1}, t_{2}, \cdot \cdot \cdot, t_{N}), \mathbf{X}=\{\mathbf{x_{1}}, \mathbf{x_{2}}, \cdot \cdot \cdot, \mathbf{x_{N}}\}.$
Taking the logrithm of the likelihood function and making use of the form of Guassian Distribution, we have
$$\log p(\mathbf{t}|\mathbf{X},\mathbf{w},\beta)=\frac{N}{2} \log \beta - \frac{N}{2} \log (2\pi) - \frac{\beta}{2} L(\mathbf{w}) ,$$ where the "square loss" is defined by
$$L(\mathbf{w})=\sum_{i=1}^{N}(t_{i}-y(\mathbf{x_{i}},\mathbf{w}))^{2}.$$
 Hence, maximizing the log-likelihood function is equivalent to minimizing the square loss $L(\mathbf{w})$, which is exactly what we do to solve ordinary regression problems in machine learning.

\subsubsection{Classification Problem}
The similar interpretation can be applied to classification problems.Suppose now we want to solve a K-class classification problem. Analogously, we can assume there is a statistical model:
$$t\sim y(\mathbf{x},\mathbf{w}),$$
where $y(\mathbf{x},\mathbf{w})$ is a neural network with output of size $1\times K$ and the sum of output's entries is $1$. And our goal is to derive the deterministic function $y$. In addition, we suppose $t$ has a K points distribution with probability $y(\mathbf{x},\mathbf{w})$ and the samples are i.i.d. drawn from this true K-point distribution. Therefore, we can write down the MLE:
$$p(\mathbf{t}|\mathbf{X},\mathbf{w},\beta)=\prod\limits_{i=1}^{N}{\prod\limits_{k=1}^{K}\hat{y}_{k}^{\mathrm{1}_{x_{i}=k}}}.$$ where $\hat{y}_{k}$ is the $k$-th entry of our predicted function $\hat{y}(\mathbf{x},\mathbf{w})$. Take the logarithm and we obtain the log-likelihood:\begin{align*}\mathrm{L}(\mathbf{t}|\mathbf{X},\mathbf{w})&=\sum\limits_{i=1}^{N}\sum\limits_{k=1}^{K}\mathrm{1}_{x_{i}=k} \log\hat{y_k}\\&=\sum\limits_{k=1}^{K} \#_{k} \log\hat{y_k}.\end{align*} where $\#_k$ denotes the number of samples belong to class k. Divide both sides by $N$ and use the law of large number, we have\begin{align*}\lim_{N \to +\infty}\dfrac{\mathrm{L}(\mathbf{t}|\mathbf{X},\mathbf{w})}{N}&=\lim_{N \to +\infty}\sum\limits_{k=1}^{K} \frac{\#_{k}}{N} \log\hat{y_k}\\&\rightarrow\sum\limits_{k=1}^{K}y_k\log\hat{y_k}.\end{align*}where $y_k$ is the $k$-th entry of the true function $y(\mathbf{x},\mathbf{w})$ and $\hat{y}_k$ is the $k$-th entry of the predicted function $\hat{y}(\mathbf{x},\mathbf{w})$. Note that $-\sum\limits_{k=1}^{K}y_k\log\hat{y_k}$ is exactly the categorical entropy loss function which is commonly used in machine learning problem for classification. Therefore, when n is large, minimizing the categorical entropy is equivalent to maximizing the MLE in this setting.
\subsection{Derivation of Natural Gradient Descent}
As stated before, neural networks have a probability interpretation. From this point of view, the optimization of objective function(loss function) with respect to parameters can be viewed as the optimization of objective function with respect to distributions. We have shown that each $\mathbf{w}$ (the values of the parameters) represents a distribution $p_{\mathbf{w}}(x)$. When we use gradient methods to optimize the objective function, we usually choose $-\nabla L$ as the descent direction, since $-\nabla L$ can be interpreted as the "steepest" descent direction in the sense that it yields the most reduction of $L$ per unit of change in $\mathbf{w}$. Here, the change in $\mathbf{w}$ is measured in standard Eucilidean norm $\| \cdot \|$. Formally ,we have
$$\frac{-\nabla L}{\|\nabla L\|}=\lim_{\epsilon \to 0}\frac{1}{\epsilon} \mathop{\arg\min}_{d:\|d\|\leq\epsilon} L(\mathbf{w}+d).$$
Notice that this formula depends on the Eucilidian metric of the parameter space, we may derive similar formula by using other metric of the parameter space. Since we have stated that the parameter space can be viewed as a space of distributions, we recall the well known metric of distribution: KL-divergence, where
$$KL(Q\|P)=\mathbb{E}_{Q} \log \frac{Q}{P}.$$
We can approximate the KL-divergence by Taylor expansion:
\begin{align*}
KL(p_{\mathbf{w}}\|p_{\mathbf{w}+ \Delta \mathbf{w}}) &\approx -\mathbb{E} [\nabla \log p_{\mathbf{w}}] \Delta \mathbf{w}-\frac{1}{2} \Delta \mathbf{w}^{T}\mathbb{E} [\nabla^{2} \log p_{\mathbf{w}}]\Delta \mathbf{w}\\
&= \frac{1}{2} \Delta \mathbf{w}^{T}\mathbb{E} [-\nabla^{2} \log p_{\mathbf{w}}]\Delta \mathbf{w}\\
&=\frac{1}{2} \Delta \mathbf{w}^{T} \mathbf{F}_{\mathbf{w}} \Delta \mathbf{w}
\end{align*}
where $\mathbf{F}_{\mathbf{w}}$ denote the Fisher Information Matrix at $\mathbf{w}$. More generally, the space of distributions is in fact a Riemannian manifold whose metric is given by the Fisher Information Matrix $\mathbf{F}_{\mathbf{w}}$. Thus, we define the Natural Gradient $\nabla_{N} L$by
$$\frac{-\nabla_{N} L}{\|\nabla_{N} L\|_{\mathbf{F}}}=\lim_{\epsilon \to 0}\frac{1}{\epsilon} \mathop{\arg\min}_{d:\|d\|_{\mathbf{F}}\leq\epsilon} L(\mathbf{w}+d),$$
where the norm$\|\cdot\|_{\mathbf{F}}$ is defined by $\| v \|_{\mathbf{F}}=\sqrt{\mathbf{v}^{T} \mathbf{F} \mathbf{v}}$ (The norm is well-defined since $\mathbf{F}$ is positive definite). We can show that
$$\nabla_{N} L=\mathbf{F}^{-1} \nabla L$$
In fact,
\begin{align*}
\lim_{\epsilon \to 0}\frac{1}{\epsilon} \mathop{\arg\min}_{d:\|d\|_{\mathbf{F}}\leq\epsilon} L(\mathbf{w}+d)&=
\lim_{\epsilon \to 0}\frac{1}{\epsilon} \mathbf{F}^{-\frac{1}{2}}\mathop{\arg\min}_{c:\|c\|\leq\epsilon} L(\mathbf{w}+\mathbf{F}^{-\frac{1}{2}}c)\\
&=\mathbf{F}^{-\frac{1}{2}} ( \frac{-\mathbf{F}^{-\frac{1}{2}}\nabla L}{\|\mathbf{F}^{-\frac{1}{2}}\nabla L\|})\\
&=\frac{-\mathbf{F}^{-1} \nabla L}{\|\mathbf{F}^{-1} \nabla L\|_{\mathbf{F}}}
\end{align*}
Thus,
$$\nabla_{N} L=\mathbf{F}^{-1} \nabla L.$$
If we choose the negative natural gradient $-\nabla_{N} L$ as the descent direction, we then derive the algorithm called "natural gradient descent":
\begin{align*}
\mathbf{w}_{n+1}&=\mathbf{w}_{n}-\eta \cdot \nabla_{N} L\\
&=\mathbf{w}_{n}-\eta \cdot \mathbf{F}^{-1} \nabla L
\end{align*}
where the Fisher Information Matrix $\mathbf{F}=\mathbb{E} [-\nabla^{2} \log p_{\mathbf{w}_{n}}(z)]=\mathbb{E} [(\nabla \log p_{\mathbf{w}_{n}}(\mathbf{z}))^{T} (\nabla \log p_{\mathbf{w}_{n}}(\mathbf{z}))].$
\subsection{Advantages of Natural Gradient Descent}
In this section, we summarize some advantages of NGD(Natural Gradient Descent).\par
First, NGD is "natural", since it optimizes the objective function in the distribution space. As we've showed before, a set of certain values of parameters are actually a representation of a distribution in our statistical model. We may reparametrized the model, then the parameters may change, but the corresponding distribution remains the same. We can say that the natural gradient reflects some geometry properties of the distribution space, for example, by using NGD, we can jump over plateaus of $p_{\mathbf{w}}(t|\mathbf{x})$. Since the loss function $L$ is a function of $p$ for certain $\{ \mathbf{(x_{n},t_{n})} \}$, plateaus of $p$ usually match those of $L$ and hence NGD can jump over plateaus of the error function too. On the other hand, NGD not only jump over plateaus, but also avoid jumping too far in each step, which may prevent overfitting to some extent. (The distributions which overfit the model are often very "rough", so in the sense of KL-divergence, they are likely far from the initial point of optimization.)\par
Another advantage is that, NGD can be viewed as a second order method, since $\mathbf{F}$ is the expectation of the Hessian of the log-likelihood function. Second order methods often perform better than first order methods, since they reveal more information about the loss function. But as we state in the introduction part, Hessian in high-dimensionality is hard to compute, which prevent us from directly using second order methods in fitting neural networks. But fortunately, though superficially second-order, NGD can in fact be written in a form of first order method.
This is derived from the special property of Fisher Information Matrix:
$$\mathbf{F}=\mathbb{E} [-\nabla^{2} \log p_{\mathbf{w}_{n}}(z)]=\mathbb{E} [(\nabla \log p_{\mathbf{w}_{n}}(\mathbf{z}))^{T} (\nabla \log p_{\mathbf{w}_{n}}(\mathbf{z}))].$$
Thus we do not need to compute the hessian, instead we compute the gradient of the log-likelihood function, and approximate the expectation by some kind of "average" of $(\nabla \log p_{\mathbf{w}_{n}}(\mathbf{z}))^{T} (\nabla \log p_{\mathbf{w}_{n}}(\mathbf{z}))$. The details will be shown in the latter parts.

\section{\textbf{AdaSqrt: A New Adaptive Optimization Algorithm}}
\subsection{From Previous Adapative Methods to AdaSqrt}
The adaptive methods, though varying from each other in certain aspects, share the same characteristic: the denominator of $\Delta \theta_{t}$ contains a sum (or average) of $g_{t}^{2}$, where $g_{t}$ denotes the gradient of $L(\theta)$. For example, in adagrad,
 $$\theta_{t+1}={\theta_{t}}-\frac{\eta g_{t}}{\sqrt{\sum_{i=1}^{t} g_{t}^{2} + \epsilon}}.$$
 The original paper\cite{Duchi} explains the intuition for adagrad, that is, $\sqrt{\sum_{i=1}^{t} g_{t}^{2} + \epsilon}$ in the denominator enforces the algorithm to descend faster in the directions where the parameter have been scarcely moved. Though reasonable, this statement cannot explain why we need to calculate the square root of $\sum_{i=1}^{t} g_{t}^{2}$. Tradition view simply claim that the square root is indeed important and updating the parameters without using it will lead to terrible training performance[an overview of ***]. Both rigorous proofs or heuristic explanations are largely lacking.\\
 In this section, we will give another explanation for $\sum_{i=1}^{t} g_{t}^{2}$, and show that the square root in the denominator is actually unnecessary under proper scaling. Recall the formula of NGD:
 $$
\mathbf{w}_{n+1}=\mathbf{w}_{n}-\eta \cdot \mathbf{F}^{-1} \nabla L
$$
where the Fisher Information Matrix $\mathbf{F}=\mathbb{E} [(\nabla \log p_{\mathbf{w}_{n}}(\mathbf{z}))^{T} (\nabla \log p_{\mathbf{w}_{n}}(\mathbf{z}))].$
Notice that for a fixed $\mathbf{z}=(\mathbf{x}_{i},{t}_{i})$, $\nabla \log p_{\mathbf{w}_{n}}(\mathbf{z})= -\nabla L(\mathbf{w}, \mathbf{x}_{i})$, thus
$$(\nabla \log p_{\mathbf{w}_{n}}(\mathbf{z}))^{T} (\nabla \log p_{\mathbf{w}_{n}}(\mathbf{z}))=(\nabla L(\mathbf{w}, \mathbf{x}_{i}))^{T} (\nabla L(\mathbf{w}, \mathbf{x}_{i})).$$
Hence, to some extent, $g_{t}^{T} g_{t}$ can be viewed as an approximation of the Fisher Matrix, which explains the occurence of $g_{t}^{2}$ in the adaptive methods, from the view of NGD. Moreover, the sqrt over $\sum_{i=1}^{t} g_{t}^{2}$ is unnecessary, since there are no sqrt over the Fisher Matrix. We will then derive a new algorithm base on this observation.
More precisely, an expectation has to be taken over $\mathbf{z}$. But in practice, we have no access to the distribution of $\mathbf{z}$. Therefore, some average of $g_{t}^{2}$ is necessary for approximating the expectation. We may approximate the Fisher Matrix by $\frac{\sum_{i=1}^{t} g_{t}^{2}}{\alpha_{t}}$, where $\alpha_{t}$ is a tuning parameter. Choosing $\alpha_{t} =\sqrt{t}$, we then derive our algorithm, AdaSqrt:
\begin{algorithm}
\caption{\textbf{AdaSqrt}}
\textbf{Require:} $\eta$: Stepsize\\
\textbf{Require:} $f(\theta)$: Stochastic objective function with parameters $\theta$ \\
\textbf{Require:} $\epsilon$ : A small constant\\
\quad$\theta_0$(Initial parameter vector)\\
\quad$S_{0}\leftarrow0$(initialize sum of square of gradients)\\
\quad$\alpha_t\leftarrow 1$(Initialize the scaling parameter)\\
\quad$t\leftarrow 0$(Initialize timestep) \\
\qquad\textbf{while} $\theta_t$ not converge \textbf{do}\\
\hspace{4em}$t\leftarrow t+1$\\
\hspace{4em}$\alpha_t\leftarrow\sqrt{t}$\\
\hspace{4em}$g_t\leftarrow\nabla_\theta f(\theta_{t-1})$\\
\hspace{4em}$S_t\leftarrow S_{t-1}+g_t^2$\\
\hspace{4em}$\theta_{t+1}\leftarrow\theta_{t}-\eta\dfrac{\alpha_{t} g_{t}}{{ S_t+\epsilon}}$\\
\qquad\textbf{end while}\\
\textbf{return} $\theta_t$(Resulting parameters)
\end{algorithm}\\
where $\eta$ is the learning rate, $\alpha_t=\sqrt{t}$, $g_t$ is the derivative of loss function at $t$-th iteration and $\epsilon$ is a small constant to ensure that the denominator is positive. Besides, in this algorithm, all updates are performed entrywisely.\par
We can see from the update formula that the denominator goes to infinity when t is large, which is similar to Adagrad. However, we know one of Adagrad's potential weakness is that the learning rate would eventually become very small due to accumulation of square of gradients. Thus, in some sense, our algorithm can be interpreted as an adaptive method adding a $\sqrt{t}$ term in numerator to counteract the rapidly increasing denominator, which would result in a more reasonable and effective learning rate. \par
In the following part we will prove a theoretical result of AdaSqrt and in section$7$ experiments are provided to examine the effect of our algorithm empirically and backup our theoretical result.
\subsection{Convergence Analysis}
We analyze the convergence of AdaSqrt under the ordinary deep learning framework, with several mild assumptions.
Given an convex objective function (loss function) $f(\theta)$ which we want to minimize, at each time $t$,our goal is to update the parameter $\theta_{t}$ using AdaSqrt, that is,
$$\theta_{t+1}=\theta_{t}-\frac{\alpha_{t} g_{t}}{\sum_{i=1}^{t}g_{t}^{2}},$$
where $g_{t}=\nabla f(\theta_{t})$, and $g_{t}^{2}$ be the elementwise square of $g_{t}$.
Since the sequence of
$\{\theta_{t}\}$ is unknown in advance, we will evaluate our algorithm using the regret $R(T)$, which is defined as
$$R(T)=\sum_{t=1}^{T}[f(\theta_{t})-f(\theta^{\ast})],$$
where $\theta^{\ast}=\mathop{\arg\min}_{\theta}f(\theta)$.
Define $G_{t}=\frac{\alpha_{t}}{\sum_{i=1}^{t}g_{t}^{2}}$. We will show that the asymptotic behavior of $G_{t}$ is crucial for convergence. If $\{G_{t}\}$ does not vary a lot for large $t$, then with $\alpha_{t}=\sqrt{t}$, AdaSqrt has a $O(\sqrt{T})$ bound. With different value of $\alpha_{t}$, the bound may be different. The theorem and related discussions will be shown in the latter part. \\

\textbf{Theorem 2 (Convergence Theorem).} 
Assume that the distance between $\theta_{t}$ and $\theta^{\ast}$ is bounded, $\|\theta_{t}-\theta^{\ast}\|_{2} \leq D, \forall \theta_{t}$ generated by AdaSqrt. Let $\alpha_{t}=\sqrt{t}$, and assume that $\{G_{t}\}$ monotonously decreases to $G\neq 0$. Then AdaSqrt achieves the following guarantee,
$$R(T)=O(\sqrt{T}) .$$\\
\textit{Proof:} Since $f: \mathbb{R}^{d} \rightarrow \mathbb{R}$ is convex,
$$f(y) \geq f(x) +\nabla f(x)^{T}(y-x).$$
Therefore,
$$f(\theta_{t})-f(\theta^{\ast})\leq g_{t}^{T} (\theta_{t}-\theta^{\ast}).$$ Then we have
\begin{align*}
R(T)&=\sum_{t=1}^{T}[f(\theta_{t})-f(\theta^{\ast})]\\
&\leq \sum_{t=1}^{T} g_{t}^{T} (\theta_{t}-\theta^{\ast})\\
&=\sum_{t=1}^{T} \sum_{i=1}^{d} g_{t,i} (\theta_{t,i}-\theta^{\ast}_{,i}) \\
&=\sum_{i=1}^{d} \sum_{t=1}^{T} g_{t,i} (\theta_{t,i}-\theta^{\ast}_{,i})
\end{align*}
where $g_{t,i} , \theta_{t,i} , \theta^{\ast}_{,i}$ be the corresponding $i^{th}$ element. Since our algorithm, AdaSqrt, is actually elementwise, we will then consider each element. For convenience, for a fixed $i$, we will omit the subscript $i$ for next several lines. Recall the update rule
$$\theta_{t+1}=\theta_{t}-\frac{\alpha_{t} g_{t}}{\sum_{i=1}^{t}g_{t}^{2}}= \theta_{t}-G_{t}g_{t}.$$Subtract the scalar $\theta^{\ast}$ and square both sides, we have

\begin{align*}
(\theta_{t+1}-\theta^{\ast})^{2}&=(\theta_{t}-\theta^{\ast}-G_{t}g_{t})^{2}\\
&=(\theta_{t}-\theta^{\ast})^{2} -2(\theta_{t}-\theta^{\ast})G_{t}g_{t} + G_{t}^{2}g_{t}^{2}
\end{align*}
Rearrage the above equation, then
$$(\theta_{t}-\theta^{\ast})g_{t} = \frac{(\theta_{t}-\theta^{\ast})^{2}-(\theta_{t+1}-\theta^{\ast})^{2}}{2G_{t}}+\frac{1}{2} G_{t}g_{t}^{2}$$
Sum the above equation for $t=1,\cdots, T$, we have
\begin{align*}
\sum_{t=1}^{T}(\theta_{t}-\theta^{\ast})g_{t} &= \sum_{t=1}^{T}\left[\frac{(\theta_{t}-\theta^{\ast})^{2}-(\theta_{t+1}-\theta^{\ast})^{2}}{2G_{t}}+\frac{1}{2} G_{t}g_{t}^{2}\right]\\
&= \frac{1}{2} \sum_{t=1}^{T} \left[(\theta_{t}-\theta^{\ast})^{2}(\frac{1}{G_{t}}-\frac{1}{G_{t-1}})\right] - \frac{1}{2} (\theta_{T+1}-\theta^{\ast})^{2} \frac{1}{G_{T}} + \frac{1}{2} \sum_{t=1}^{T} G_{t}g_{t}^{2}\\
&\leq  \frac{1}{2} \sum_{t=1}^{T} \left[D^{2} (\frac{1}{G_{t}}-\frac{1}{G_{t-1}})\right] + \frac{1}{2} \sum_{t=1}^{T} G_{t}g_{t}^{2}  \\
&= \frac{1}{2} D^{2}\frac{1}{G_{T}} + \frac{1}{2} \sum_{t=1}^{T} G_{t}g_{t}^{2}
\end{align*}
When $T \rightarrow \infty$, since $\{G_{t}\}$ converges to $G\neq 0$ , the first term converges to a constant. For the second term, since $G_{t}$
monotonously decreases,
\begin{align*}
\frac{1}{2} \sum_{t=1}^{T} G_{t}g_{t}^{2} &\leq \frac{1}{2} G_{1} \sum_{t=1}^{T} g_{t}^{2}\\
&=\frac{1}{2} G_{1} \frac{\alpha_{T}}{G_{T}}\\
&=\frac{1}{2} G_{1} \frac{\sqrt{T}}{G_{T}}\\
&=O(\sqrt{T})
\end{align*}
Hence we've shown that
$$\sum_{t=1}^{T} g_{t,i} (\theta_{t,i}-\theta_{,i}^{\ast})=O(\sqrt{T}), \forall i=1,\cdots,d.$$
Thus $$R(T) \leq \sum_{i=1}^{d} \sum_{t=1}^{T} g_{t,i} (\theta_{t,i}-\theta^{\ast}_{,i})=O(\sqrt{T}),$$
which completes the proof. \hspace{30em}$\qedsymbol$\\

We can deduce from the proof that different $\alpha_{t}$ will lead to different upper bounds. It seems that if we choose a smaller $\alpha_{t}$, which means a larger learning rate, we will get a tighter bound. But the crucial point is that, we need the assumptions to be achieved. In training, we usually expect the actual learning rate $\|G_{t}\|$ to decrease in order for the convergence of algorithm. Intuitively, $\|G_{t}\|$ should not vary a lot, which can be seen from our proof. Some related works have shown that adaptive methods such as Adam, occasionally use extreme learning rate--too large or too small in the late phase of training \cite{1902.09843}, which results in poor performance. However, our algorithm seeks to maintain a moderate learning rate according to the experiments. We believe that the tuning parameter $\alpha_{t}$ shows some information of the geometry of the loss surface, and we leave this to future works.

\section{Experimental Results}
We perform our experiments on two real-world datasets MNIST and CIFAR-10. In experiments, we compare our method and other first-order optimization methods like SGD, Adagrad and Adam's performance under certain neural network structure including multi-layer fully-connected neural networks and convolutional neural networks. We use same batches of samples for all these methods and evaluate the training loss and prediction error after training. The step sizes for these methods are chosen such that each algorithm have relatively optimal performance in experiments.\par
Using these complicated models and large datasets, we demonstrate that our method can efficiently solve certain practical deep learning problems. Furthermore, our method even outperforms others under certain neural network structure for real problems.
\subsection{Experiments on MNIST}
 MNIST is a real database containing a training set of $60000$ black and white handwritten digits as well as a test set of $10000$ handwritten digits. Each image has a size of $28\times28$ and a corresponding label which indicates the true number in the image. Our task is to use the training set to fit a neural network such that the prediction error is as small as possible. For this ten-classification problem, we use sparse categorical entropy as loss function$\text{i.e. loss}=\sum_{i=1}^{n} y_{i} \log \left(\hat{y}_{i}\right)$ where $y$ is the ground truth and $\hat{y}$ is the predictive probability. In experiments, we construct a simple two-layer fully connected neural network with hidden layers of 300 units and 100 units respectively. We use $ReLU$ activation in both layers and a minibatch size of $50$ for training. After choosing optimal step sizes for each methods, we train the model using these optimization methods with same batches of samples and compare the training loss and prediction error after several epochs. The result are shown as follows:
\begin{figure}[htb]
\includegraphics[width=23em]{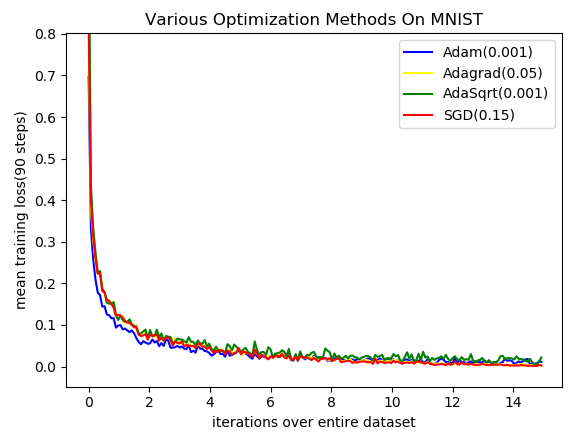}
\includegraphics[width=23em]{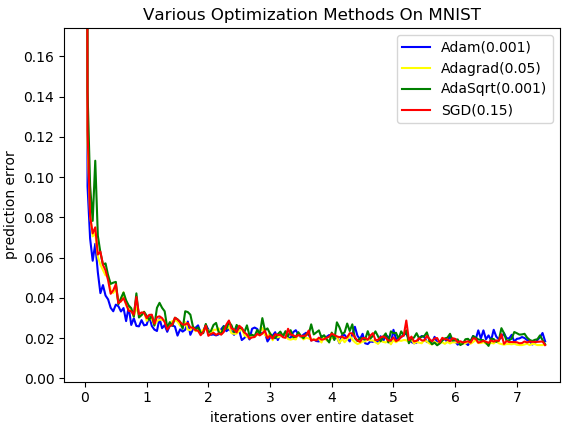}
\end{figure}
 It is easy to see that Adam rapidly reduces the training loss and prediction error at the beginning of training(first three epochs) while prediction error of others oscillate relatively larger at first. However, when the predicted model is close to the real one(after five epochs), Adam would suffer from high variance due to the non-convergent learning rate. On the other hand, Adagrad's learning rate would converge to zero and therefore the loss and error are more stable than other methods. Furthermore, although SGD converges relatively slow at first, its strong ability of exploring the local structure makes it more powerful when the model is near the minimum.\par As for the proposed method, we can see that it has the best performance in $9\sim13$ epochs. Since the parameters are relatively easy to converge, the sum of square of gradients in denominator of the learning-rate term is sometimes not large enough to counteract the $\sqrt n$ in numerator, which results in a relatively large step size and high variance in the final few epochs. A natural way improve our algorithm is to set up a threshold w.r.t. the gradients and use it to fix a $n_0$ in the numerator or reduce the exponential coefficient when $n$ is large.\par
 However, although our method might have certain drawbacks in this problem, it still has almost the same effect as other optimization methods in this practical problem such as less than two percentages' prediction error, rapid convergence and low computational cost. To some extent, our adaptive method can be applied as an alternative for certain optimization problems.
\subsection{Experiments on CIFAR-10}
We then compare our method with other optimization methods on CIFAR-10. CIFAR-10 is a dataset containing a training set of $50000$ $32\times 32$ color images and a testing set of$10000$ $32\times32$ color images. All these images belong to one of ten classes and are labeled with corresponding digits. Our goal is to minimize the ten-class classification problem's predictive error through learning from training set. Although our task is similar to previous section's, training on CIFAR-10 is far more difficult than on MNIST since those color images have more intricate local structure handwritten digits. Therefore, we construct deep convolutional neural networks for this problem.\par
In the problem, we construct a neural network with 3 convolutional layers each of which followed by a Maxpooling layer respectively and a fully connected layer with 128 hidden units. Due to the intrinsic complexity of this classification problem, we can only achieve approximately $80\%$ predictive accuracy after $15$ iterations over the entire dataset. However, this would provide us a better chance to compare various optimization methods' performance since the model is hard to converge. In the following experiments, we train the model with a minibatch of size 100 and run $7500$ iterations such that we can approximately cover the training set $15$ times. Following is our result:\par
\begin{figure}[htb]
\includegraphics[width=23em]{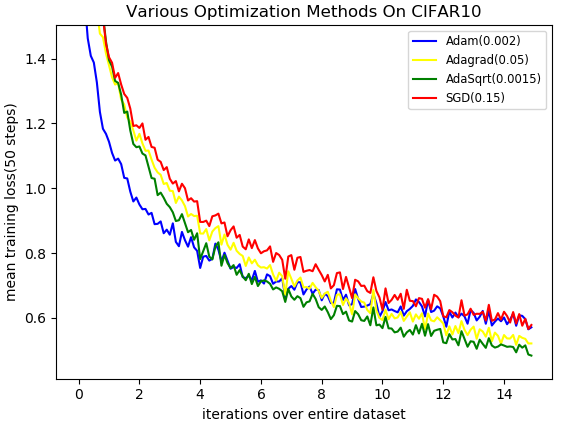}
\includegraphics[width=23em]{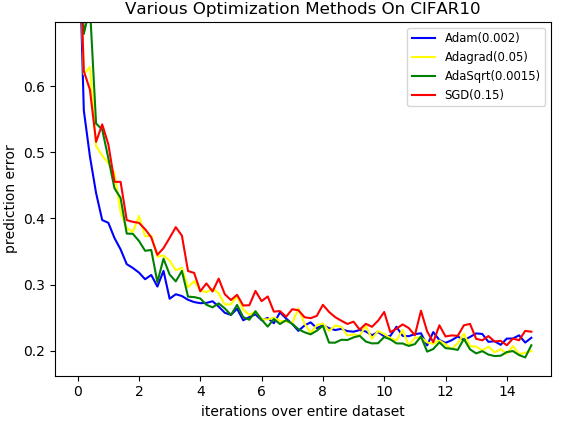}
\end{figure}
It is clear from the graphic that in the first 5 epochs, Adam reduces the training loss faster than other three methods. However, after that, our method has similar average training loss like Adam and finally outperforms Adam and becomes the best method among them to reduce both the training loss and the predictive error. Furthermore, we can see that our method always have better performance than Adagrad and SGD with fixed step size during the entire training process. Since our algorithm is adapted from Adagrad, we can conclude that our method indeed have better performance than the original version in this practical problem and has similar or even better performance than other powerful method like Adam in late epochs. \par
How can we improve our AdaSqrt algorithm's performance in first few epochs? Although we have proved that the exponential coefficient $\frac{1}{2}$ is optimal in some sense, we can also use $0.6,0.7,0.9,1$ or other suitable real number as alternatives. These replacements can accelerate our algorithm at first but would result in high variance. Therefore, a probably feasible method is to use different exponential coefficients in the training procedure(large at first and small later). For lacking of time, we leave the design of such algorithms to future work.\par
In conclusion, based on our algorithm's efficient performance, it is convincing for us to claim that the $\sqrt{} $ in the denominator of Adam, Adagrad and other adaptive methods are indeed not necessary. In order to obtain an efficient first order optimization algorithm, it suffices to let the algorithm taking advantage of second order information of the loss function and the use of square root can be compensated by proper scaling. Although there is an interesting observation that taking the square root can make those adaptive methods homogeneous and the parameters' update process would therefore keep invariant under scaling of loss function, we can overcome the defect of non-homogeneity by tuning the learning rate manually like SGD, Nestrov. Furthermore, the learning rate in our algorithm is less likely to explode since the denominator would increase rapidly when some gradients are large and we can also fix a $n_0$ as numerator when the model almost converges.

\section{Conclusions and Future works}
In this article, we mainly focus on uncovering second-order information in first-order optimization algorithms. For momentum, we rigorously prove that Nesterov Accelerated Gradient explores second-order information by using the difference of gradients while training hence is better than the original Momentum. For adaptive methods, we rigorously prove that Adagrad and Adam can be regarded as relaxations of Natural Gradient Descent, a well-known second-order technique in computation statistics, only with a slight difference in the square root of denominator. Based on this observation, we design a new algorithm, AdaSqrt, which outperforms Adam on MNIST and CIFAR10. Our algorithm demonstrates that the tradition view concerning the importance of square root might be completely wrong---with proper scaling, $\sqrt{T}$, we can achieve even better performance.

We strongly believe that going beyond gradient is a promising direction with huge practical value that deserves more attention in the future. Of course, by claiming this, we are referring to methods without directly computing Hessian. To the best of our knowledge, there are at least four possible means to achieve this goal. The first approach is to use some technical tricks in matrix operation---which is exactly what BFGS algorithm does---to approximate the Hessian. The second approach is to use the idea of Newton method or Natural Gradient Descent, which motivates the design of AdaSqrt as in this paper. The third approach is to mimic the insight of Nesterov Accelerated Gradient, which predicts the location of one-step future by using past gradients. We notice that a long-run prediction might also be made, though in certain sacrifice of accuracy, and starting from a two-steps future prediction sounds plausible. In such case, the new algorithm actually explores higher-order information, which further accelerates the training. Due to the limit of time, we leave the design of such algorithm to future work. The last approach is probably the most difficult but also the most intriguing one: we encourage the readers to uncover high-dimensional geometry properties of the energy landscape of deep neural networks. So long as we have a better understanding of the loss surface (for example, the loss surface might be a union of low-dimensional manifolds), we can design provable algorithms that have both faster training speed and better generalization performance.

\section*{Acknowledgements}
The authors are partially supported by the elite undergraduate training program of School of Mathematical Sciences in Peking University. They want to thank Huiyuan Wang for his helpful discussion and Putian Li for his suggestion in typesetting.
\quad\\
\quad\\

\end{document}